\documentclass[letterpaper, 10 pt, conference]{ieeeconf}  

\usepackage{amssymb}
\usepackage{amsmath}
\usepackage{commath}
\usepackage{mathtools}

\usepackage{import}
\usepackage{paralist}
\usepackage{svg}
\usepackage{textcomp}
\usepackage{gensymb}
\usepackage{dsfont}
\usepackage{subfig}
\usepackage{upgreek}
\usepackage[ruled]{algorithm2e}
\usepackage{bm}

\IEEEoverridecommandlockouts                              

\usepackage{graphics} 
\usepackage{epsfig} 
\usepackage{todonotes}
\usepackage[normalem]{ulem}
\usepackage{hyperref}

\title{\LARGE \bf
Non-Linear Trajectory Optimization for Large Step-Ups: \\
Application to the Humanoid Robot Atlas
}

\author{Stefano Dafarra$^{1}$, Sylvain Bertrand$^{2}$, Robert J. Griffin$^{2}$, Giorgio Metta$^{1}$, Daniele Pucci$^{1}$, Jerry Pratt$^{2}$ 
\thanks{${}^{1}$ iCub Facility Department, Istituto Italiano di Tecnologia, 16163 Genova,
Italy (e-mail: \texttt{name.surname@iit.it})}%
\thanks{${}^{2}$ Florida Institute for Human and Machine
Cognition, IHMC, 40 South Alcaniz Street, Pensacola, Florida 32502, United States (e-mail: \texttt{\{sbertrand,rgriffin,jpratt\}@ihmc.us})}%
}

\DeclareMathOperator*{\minimize}{minimize}
\DeclareMathOperator*{\minimizebf}{\textbf{minimize}}

\newcommand\Tstrut{\rule{0pt}{2.6ex}}         
\newcommand\Bstrut{\rule[-0.9ex]{0pt}{0pt}}   

\begin{document}
	
\maketitle
\thispagestyle{empty}
\pagestyle{empty}

\begin{abstract}
Performing large step-ups is a challenging task for a humanoid robot. It requires the robot to perform motions at the limit of its reachable workspace while straining to move its body upon the obstacle.	
This paper presents a non-linear trajectory optimization method for generating step-up motions. We adopt a simplified model of the centroidal dynamics to generate feasible Center of Mass trajectories aimed at reducing the torques required for the step-up motion. The activation and deactivation of contacts at both feet are considered explicitly.
The output of the planner is a Center of Mass trajectory plus an optimal duration for each walking phase. These desired values are stabilized by a whole-body controller that determines a set of desired joint torques.
We experimentally demonstrate that by using trajectory optimization techniques, the maximum torque required to the full-size humanoid robot Atlas can be reduced up to 20$\%$ when performing a step-up motion. 

\end{abstract}

\section{Introduction}

A fundamental ability of humanoid robots consists in moving over steps and stairs, where they can exploit the legged configuration. A possible approach for the generation of humanoid motions over stairs is to tackle the problem as an extension of the planar walking motion generation \cite{hirai1998development,hu2016walking, caron2019stair}. This paper presents methods and validations for extending the planar walking generation algorithms to large step-up behaviors of humanoid robots.

Assuming the angular momentum to be constant and restricting the robot center of mass (CoM) on a plane at fixed height, it is possible to derive simple and effective control laws based on the well known Linear Inverted Pendulum (LIP) model \cite{kajita20013d}. By exploiting the linearity of the model, it is possible to determine analytical stability conditions, enabling for reactive motions in case of disturbances \cite{Pratt2006,griffin2017walking}. However, the linearity of the model is based on the assumption of constant CoM height, resulting in a fixed pendulum constant $\omega$. Instead, when considering $\omega$ as a tuning parameter, the model linearity can be preserved even in case of varying CoM height  \cite{englsberger2013three}. If it is left unconstrained, it is possible to generate walking trajectories with a straight knee configuration \cite{griffin2018straight}.
In other works, balance is maintained by varying only the CoM height \cite{koolen2016balance}, hence completely removing the assumption for it to be constant. The robustness properties of this strategy have been further studied through the use of Sums-of-Squares optimization \cite{posa2017balancing}. A similar model has been used also in a multi-contact scenario \cite{perrin2018effective} and extended considering zero-moment point (ZMP) \cite{vukobratovic2004zero} motions to determine stability conditions \cite{caron2017dynamic,caron2018balance}. These approaches have been used to control a humanoid robot considering only single support phases. In addition, the CoM height variations still remain very limited.

\begin{figure}[tpb]
    \centering
    \includegraphics[width=0.75\columnwidth]{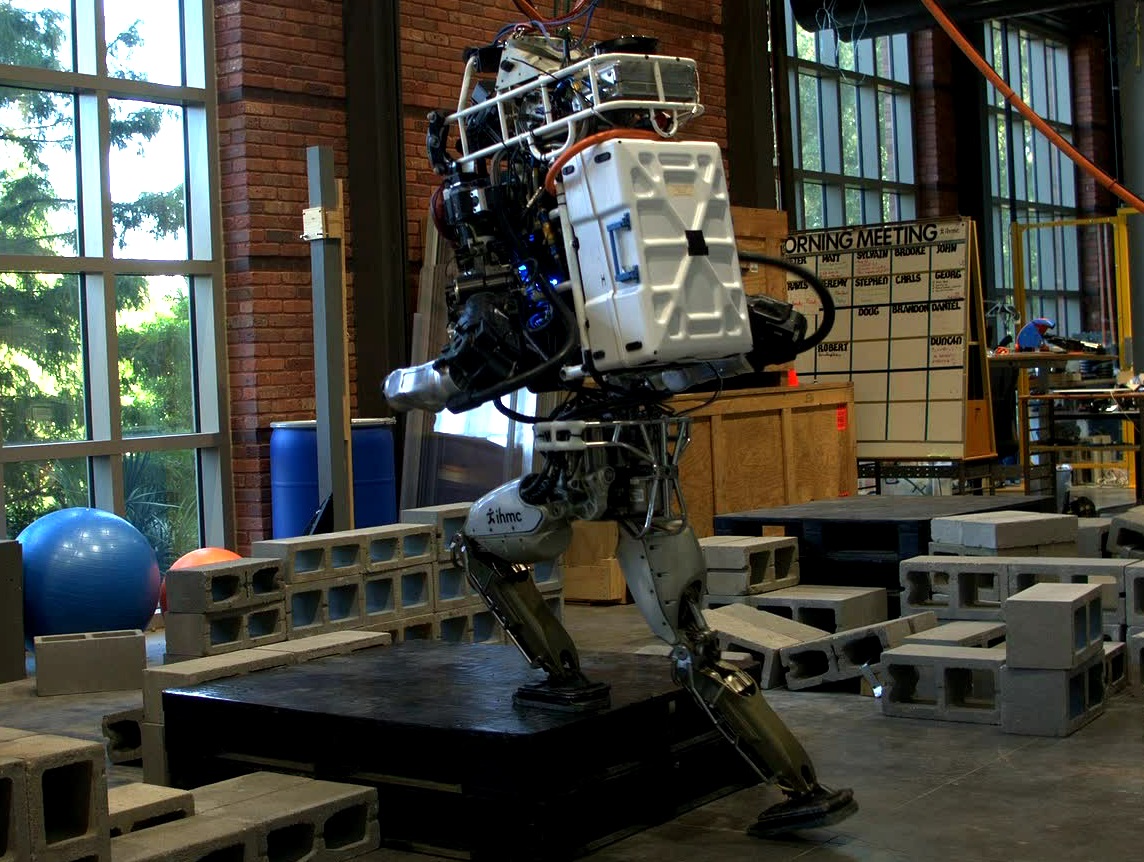}
    \caption{Atlas performing a 31$\mathrm{cm}$ tall step-up.}
    \label{fig:atlasShowingOff}
\end{figure}

In case of large step-ups, the significant and required variation of the CoM height is no more negligible and it cannot be considered as constant. In addition, it becomes important to perform such motion dynamically, exploiting the robot momentum to avoid excessive effort from the robot actuators.
In literature, this problem is faced by using non-linear trajectory optimization techniques. Apart from the problems arising when using non-linear, and potentially non-convex, models, an additional hurdle is the determination of contact timings and locations. For example, kinematic and dynamic quantities can be computed in an iterative fashion, assuming the contact sequence to be known \cite{herzog2015trajectory}. Authors of \cite{winkler18} adopt a phase-based approach, where the number of contact phases are predetermined, but their duration and the contact location are a result of the optimization problem. Integer variables can be used to describe the activation/deactivation of contacts \cite{aceituno2018simultaneous}.
Other contributions include complementarity conditions as constraints \cite{dai2014whole, posa2014direct}, thus contact locations and timings are an output of the optimization problem. The time needed to find a solution ranges from seconds to several minutes, while results are shown only in simulation. On the other hand, methods that rely on iterative Linear Quadratic Regulators usually require small computational times and are successfully applied to robots with point contacts, like quadrupeds or a single leg hopping robot \cite{neunert2018whole,carius2018trajectory}.

The aim of this paper is to leverage trajectory optimization techniques to generate motions which allow a humanoid robot to perform large step-ups. In particular, we exploit a reduced model of the centroidal dynamics, where the forces on both feet are considered. The angular momentum is assumed constant. By adopting a \emph{phase-based trajectory optimization} technique, we plan over a fixed series of contacts. Through a particular heuristic, we are able to avoid generating trajectories which would require a high torque expenditure. Real world experiments are performed on the Atlas humanoid robot, showing a reduction of the maximal knee torque up to 20\%. 


\section{Background} \label{sec_ls: background}
\subsection{Notation}
This paper uses the following notation.
\begin{itemize}
	\item The $i_{th}$ component of a vector $\bm{x}$ is denoted as $x_i$.	
    \item The transpose operator is denoted by $(\cdot)^{\top}$.
	\item $\mathcal{I}$ is a fixed inertial frame with respect to (w.r.t.) 
	which the robot's absolute pose is measured. Its $z$ axis is supposed to point against gravity, while the $x$ direction defines the forward direction.
	\item Given a function of time $f(t)$ the dot notation denotes the time derivative, i.e.
	$\dot{f} := \frac{\dif f}{\dif t}$. Higher order derivatives are denoted with a corresponding amount of dots.
 	\item $0_{n \times n} \in \mathbb{R}^{n \times n}, n \in \mathbb{N}^+,$ denotes a zero matrix.
 	\item The weighted L2-norm of a vector $\bm{v} \in \mathbb{R}^n$ is denoted by $\|\bm{v}\|_W$, where $W \in \mathbb{R}^{n\times n}$ is a weight matrix.
	\item $^{A}\bm{R}_{B} \in SO(3)$ and $^{A}\bm{H}_{B} \in SE(3)$ denote the rotation and transformation matrices which transform a vector expressed in the $B$ frame, $^B \bm{x}$, into a vector expressed in the $A$ frame, $^A \bm{x}$.
\end{itemize}

\subsection{QP-based whole-body controller} \label{sec_ls:qp_controller}
The robot motion is controlled by means of a QP-based whole-body controller, presented in \cite{koolen2016design}. It determines a set of desired joint accelerations, including the spatial acceleration of the floating-base joint with respect to $\mathcal{I}$. We denote this quantity with $\dot{\bm{\nu}}_d \in \mathbb{R}^{n+6}$ where $n$ is the number of joints under control. In addition, the whole-body controller generates desired contact wrenches.

The controller handles different motion tasks, amongst which Cartesian and center of mass (CoM) tasks.
The former are achieved by minimizing the following quantity:
\begin{equation}
    \left\|\bm{J}_t \dot{\bm{\nu}}_d - \left(\bm{\alpha}_d - \dot{\bm{J}}_t \bm{\nu}\right) \right\|^2,
\end{equation}
where $\bm{J}_t \in \mathbb{R}^{n_j \times (n+6)}$, is the task Jacobian, with $n_j$ being the number of degrees of freedom of task $j$. $\bm{\alpha}_d \in \mathbb{R}^{n_j}$ is the task desired acceleration, while $\bm{\nu}$ are the \emph{measured} base and joint velocities (obtained through an internal estimator). The quantity $\left(\bm{\alpha}_d - \dot{\bm{J}}_t \bm{\nu}\right)$ is a vector which does not depend on the decision variables and it is referred with the symbol $\bm{\beta}_i$.

The CoM task is achieved by controlling the robot momentum $\bm{h} \in \mathbb{R}^6$. In particular, we exploit the following relation:
\begin{equation}
    \dot{\bm{h}} = \bm{J}_\text{CMM} \dot{\bm{\nu}} + \dot{\bm{J}}_\text{CMM} \bm{\nu} = \bm{W}_g + \sum_i \bm{W}_{\text{gr}, i} + \sum_i \bm{W}_{\text{ext}, i},
\end{equation}
where $\bm{J}_\text{CMM} \in \mathbb{R}^{6 \times (n+6)}$ is the Centroidal Momentum Matrix \cite{orin08}, $\bm{W}_g = \left[ 0_{3\times 1}^\top ~  m\bm{g}^\top\right]^\top$ is the wrench due to gravity. $\sum_i \bm{W}_{\text{ext}, i}$ represents the summation of external wrenches applied to the robot, except those exerted on the robot body as a result of a contact, i.e. $\sum_i \bm{W}_{\text{gr}, i}$.

A different point force is considered applied at each vertex of a contact patch. In particular, these forces are defined through a coordinate vector $\bm{\rho}_i \in \mathbb{R}^m$, whose basis are $m$ extreme rays of the friction cone. By constraining each component of $\bm{\rho}_i$ in the range $\left[\rho^\text{min}, \rho^\text{max} \right]$, we avoid excessively high wrenches. More importantly, friction and Center of Pressure \cite{sardain2004forces} (CoP) constraints are implicitly satisfied \cite{pollard2001animation}. Trivially, also the resulting normal force is bounded to be positive. It is possible to express the corresponding wrench in the inertial frame $\mathcal{I}$, $\bm{W}_{\text{gr}, i}$, as a function of $\bm{\rho}_i$:
\begin{equation}
    \bm{W}_{\text{gr}, i} = \bm{Q}_i \bm{\rho}_i,
\end{equation}
where $\bm{Q}_i \in \mathbb{R}^{6\times m}$ depends on the position of contact vertex $i$ and on the basis vectors. By concatenating all the $\bm{Q}_i$ and $\bm{\rho}_i$, we obtain $\sum_i \bm{W}_{\text{gr}, i} = \bm{Q} \bm{\rho}$. 

The Cartesian and CoM task are grouped as follows:
\begin{equation*}
        \hat{\bm{J}} = \begin{bmatrix}
                            \bm{J} \\ 
                            \bm{J}_\text{CMM}
                        \end{bmatrix}, \quad
        \hat{\bm{\beta}} = \begin{bmatrix}
                            \bm{\beta} \\ 
                            \dot{\bm{J}}_\text{CMM} \bm{\nu} - \dot{\bm{h}}_d
                   \end{bmatrix},
\end{equation*}
with $\bm{J}$ and $\bm{\beta}$ being the concatenation of all the tasks Jacobians $\bm{J}_i$ and $\bm{\beta_i}$ respectively. $\dot{\bm{h}}_d$ is a desired momentum rate of change.

At every control cycle, the solved QP problem writes:
\begin{IEEEeqnarray*}{LL}
\minimize_{\dot{\bm{\nu}}_d,~ \bm{\rho}} &\|\hat{\bm{J}} \dot{\bm{\nu}}_d - \hat{\bm{\beta}}\|^2_{C_t} + \|\bm{\rho}\|^2_{C_\rho} + \|\dot{\bm{\nu}_d}\|^2_{C_{\dot{\bm{\nu}}}} \\
\text{subject to: } &\bm{J}_\text{CMM} \dot{\bm{\nu}}_d + \dot{\bm{J}}_\text{CMM} \bm{\nu} = \bm{W}_g + \bm{Q} \bm{\rho} + \sum_i \bm{W}_{\text{ext}, i} \\
&\bm{\rho}^\text{min} \leq \bm{\rho} \leq \bm{\rho}^\text{max}.
\end{IEEEeqnarray*}
The desired joint accelerations and contact wrenches are used to compute a set of desired joint torques to be commanded to the robot.

\subsection{The variable height double pendulum}
The linear momentum rate of change can be written according to Newton's second law:
\begin{equation}\label{eq:com_point_dynamics}
    m \ddot{\bm{x}} = -m  \bm{g} + \sum_i {^\mathcal{I} \bm{f}_i},
\end{equation}
where $\bm{x} \in \mathbb{R}^3$ is the center of mass position expressed in the inertial frame $\mathcal{I}$. $m$ is the total mass of the robot, $\bm{g}$ is the gravity acceleration while $^\mathcal{I}\bm{f}_i \in \mathbb{R}^3$ is an external force expressed in $\mathcal{I}$.
We can express the wrenches applied at the feet in a frame parallel to $\mathcal{I}$, having the origin coincident with the corresponding CoP. We assume that no moment is exerted along the axis perpendicular to the foot, passing through the CoP. Thanks to these two choices, the moments are null. It is possible to enforce the angular momentum to be constant by constraining these forces along a line passing through the CoM, i.e.:
\begin{equation}\label{eq:simplified_force}
    ^\mathcal{I} \bm{f}_* = m \lambda_* \left(\bm{x} - ^{\mathcal{I}}\bm{p}_* \right).
\end{equation}
The symbol $*$ is used as a placeholder for either the left $l$ or the right $r$ foot. $^\mathcal{I}  \bm{f}_* \in \mathbb{R}^3$ is the force applied on the foot, measured in $\mathcal{I}$ coordinates. $\lambda_* \in \mathbb{R}$ is a multiplier. Since the contacts are considered unilateral, $\lambda_*$ is constrained to be greater than zero for guaranteeing the positivity of normal forces. $^{\mathcal{I}}\bm{p}_* \in \mathbb{R}^3$ is the position of the foot CoP expressed in $\mathcal{I}$, in particular:
\begin{equation}\label{eq:cop_in_foot}
    ^{\mathcal{I}}\bm{p}_* = \bm{x}_* + ^\mathcal{I}\bm{R}_* \bm{p}_*,
\end{equation}
where $\bm{x}_* \in \mathbb{R}^3$ is the foot position. $\bm{p}_*$ is the CoP expressed in foot coordinates, thus having the $z$ coordinate equal to zero. For the sake of simplicity, from now on we will drop the $\mathcal{I}$ subscript in front of the rotation matrix $^\mathcal{I}\bm{R}_*$, which expresses the relative rotation between the inertial frame $\mathcal{I}$ and a frame attached to the foot.

Assuming the forces exerted on the feet to be the only external forces, we can rewrite Eq. \eqref{eq:com_point_dynamics} as:
\begin{equation}
    \begin{split}
        m \ddot{\bm{x}} = &-m  \bm{g} + m \lambda_l \left(\bm{x} - \bm{x}_l - \bm{R}_l ~ \bm{p}_l \right) \\
        &+ m \lambda_r \left(\bm{x} - \bm{x}_r - \bm{R}_r ~ \bm{p}_r \right).
    \end{split}
\end{equation}
Dividing by the total mass of the robot, we can finally write:
\begin{equation}\label{eq:double_pendulum}
    \begin{split}
        \ddot{x} = &-\bm{g} + \lambda_l \left(\bm{x} - \bm{x}_l - \bm{R}_l ~ \bm{p}_l \right) \\
        &+ \lambda_r \left(\bm{x} - \bm{x}_r - \bm{R}_r ~ \bm{p}_r \right).
    \end{split}
\end{equation}
\section{Dynamic Planning for Large Step-Ups} \label{sec_ls:sup}
The step-up planner presented in this paper leverages the idea of \emph{phase-based trajectory optimization} explored in \cite{winkler18,carpentier2016versatile,caron2017make}. We assume to know the contact sequence beforehand, simplifying the handling of their activation and deactivation.
This section presents constraints, tasks and methodologies used to solve the corresponding non-linear trajectory optimization problem. 

We start by simplifying Eq. \eqref{eq:double_pendulum} as follows:
\begin{equation}\label{eq:dynamics_simple}
    \ddot{\bm{x}}  = \bm{a}_i.
\end{equation}
$\bm{a} \in \mathbb{R}^3$ assumes different value depending on the contact state of the corresponding phase $i$:
\begin{itemize}
    \item no contacts \begin{equation*}
        \bm{a}_i = -\bm{g};
    \end{equation*}
    \item one foot in contact ($*$ is either $l$ or $r$) \begin{equation*}
        \bm{a}_i = -\bm{g} + \lambda_* \left(\bm{x} - \bm{x}_* - \bm{R}_* ~ \bm{p}_* \right);
    \end{equation*}
    \item both feet in contact \begin{equation*}
        \bm{a}_i = -\bm{g} + \lambda_l \left(\bm{x} {-} \bm{x}_l {-} \bm{R}_l ~ \bm{p}_l \right) + \lambda_r \left(\bm{x} {-} \bm{x}_r {-} \bm{R}_r ~ \bm{p}_r \right).
    \end{equation*}
\end{itemize}
We can obtain an approximated discrete form of Eq. \eqref{eq:dynamics_simple} through a second order Taylor expansion: 
\begin{IEEEeqnarray}{RCL}
    \IEEEyesnumber \phantomsection \label{eq:discrete_double_pendulum}
    \bm{x}(k+1) &=& \bm{x}(k) +  \bm{v}(k)\mathrm{dt}_i + \frac{1}{2}\bm{a}_i(k)\mathrm{dt}_i^2 \IEEEyessubnumber\\
    \bm{v}(k+1) &=& \bm{v}(k) + \bm{a}_i(k)\mathrm{dt}_i, \IEEEyessubnumber
\end{IEEEeqnarray}
where $k \in \mathbb{R}$ is used to indicate a generic discrete instant, while $\bm{v}(k) \in \mathbb{R}^3$ is the CoM velocity at instant $k$. The duration $T_i \in \mathbb{R}$ of each phase $i$ is not defined a-priori, but it is considered as an optimization variable. Nevertheless, each phase consists of a fixed number of instants, thus $\mathrm{dt}_i = T_i/N$, where $N \in \mathbb{N}^+$ is the number of instants per phase. In addition, each $T_i$ is bounded in $\left[T_{i}^\text{ min},~ T_{i}^\text{ max}\right]$.

\subsection{Force and leg constraints}\label{sec:force_constraints}
The forces applied at the feet need to satisfy some conditions in order to be attainable by the robot. These are embedded as constraints in the optimization problem.

First of all, the CoP has to lie inside the foot polygon. This can be obtained by specifying a set of linear inequalities:
\begin{equation}
    \bm{A}_* \bm{p}_* \leq \bm{b}_*,
\end{equation}
where the matrix $\bm{A}_* \in \mathbb{R}^{v \times 3}$ and the vector $\bm{b}_* \in \mathbb{R}^{v}$ can be computed according to the location of the foot vertices, whose number is $v$. Since $\bm{p}_*$ is defined in foot coordinates, these quantities do not depend on the foot pose.

The contact forces are supposed to lie within the friction cone in order to avoid slipping. This is imposed by
\begin{equation}\label{eq:friction_in_foot}
    \sqrt{{}^*f_{*, x}^2 + {}^*f_{*, y}^2} \leq \mu_s ~ {}^*f_{*, z}.
\end{equation}
 $\mu_s \in \mathbb{R}$ is the static friction coefficient and ${}^*\bm{f}_{*} \in \mathbb{R}^3$ is the force exerted on a foot, expressed in foot coordinates, namely ${}^*\bm{f}_{*} = \bm{R}_*^\top {^\mathcal{I} \bm{f}_*}$. We assume the foot frame to be parallel to a frame attached to the ground, with the $z-$ axis perpendicular to the walking surface. Using Eq. \eqref{eq:simplified_force} and \eqref{eq:cop_in_foot}, we can rewrite Eq. \eqref{eq:friction_in_foot} as follows:
\begin{equation} \label{eq:friction_full}
    \begin{bmatrix}
        1 & 1 & -\mu_s^2
    \end{bmatrix} \left(\bm{R}_*^\top  m \lambda_* \left(\bm{x} - \bm{x}_* - \bm{R}_* ~ \bm{p}_* \right) \right)^2 \leq 0.
\end{equation}
Notice that when $\lambda_* = 0$, friction is automatically satisfied as every component goes to zero. Thus, we can simplify Eq. \eqref{eq:friction_full} as follows:
\begin{equation}
    \begin{bmatrix}
        1 & 1 & -\mu_s^2
    \end{bmatrix} \left(\bm{R}_*^\top \left(\bm{x} - \bm{x}_* - \bm{R}_* ~ \bm{p}_* \right)\right)^2 \leq 0.
\end{equation}

In order to consider the torsional friction, we impose the equivalent normal torque $\tau_{*, z}$, applied at the origin of the foot frame, to be bounded, i.e. $-\mu_t {}^*f_{*, z} \leq \tau_{*, z}\leq \mu_t {}^*f_{*, z}$, with $\mu_t \in \mathbb{R}$ the torsional friction coefficient. Given that the external force ${}^*\bm{f}_{*}$ is applied in $\bm{p}_*$, we can rewrite this constraint as follows:
\begin{equation}
    - \begin{bmatrix}
    0 \\ 0 \\ \mu_t
    \end{bmatrix}^\top {}^*\bm{f}_{*} \leq \begin{bmatrix}
                                     -p_{*,y} \\ p_{*,x} \\ 0
                                  \end{bmatrix}^\top {}^*\bm{f}_{*} \leq \begin{bmatrix}
                                                                        0 \\ 0 \\ \mu_t
                                                                    \end{bmatrix}^\top {}^*\bm{f}_{*}
\end{equation}
or equivalently
\begin{equation}
    -\bm{d}_*^\top {}^*\bm{f}_{*} \leq \bm{c}_*^\top {}^*\bm{f}_{*} \leq \bm{d}_*^\top {}^*\bm{f}_{*}.
\end{equation}
Notice that $\bm{c}_*^\top {}^*\bm{f}_{*}$ corresponds to the $z$ component of the cross product $\bm{p}_* \times {}^*\bm{f}_{*} $.
By substituting ${}^*\bm{f}_{*}$ as for the static friction constraint, we write the following set of constraints:
\begin{equation}
     \bm{F}_* \bm{R}_*^\top \left(\bm{x} - \bm{x}_* - \bm{R}_* ~ \bm{p}_* \right) \leq 0
\end{equation}
where
\begin{equation}
    \bm{F}_* = \begin{bmatrix}
                \left( \bm{c}_* - \bm{d}_*\right)^\top \\
                \left( -\bm{c}_* - \bm{d}_*\right)^\top
           \end{bmatrix}.
\end{equation}

When planning the robot motion, we have to take into consideration its kinematic limits. We can approximate the robot leg length with the distance between the CoM and the corresponding foot position, thus limiting excessively wide motions or movements which could cause a leg to collapse. This can be achieved with the following constraint:
\begin{equation}
    -l^\text{min} \leq \|\bm{x} - \bm{x}_*\| \leq l^\text{max}
\end{equation}
with $l^\text{min}, l^\text{max} \in \mathbb{R}$ the lower and upper bound, respectively. These constraints can be applied on each leg separately.

\subsection{Tasks} \label{sec:double_pendulum_tasks}
The cost function is designed to generate a CoM trajectory for the robot to perform large step-ups exploiting its momentum. It is composed of several tasks. The full cost function is shown at the top of Optimization Problem \ref{alg:step-up-planner}.

The first task weights the distance of the terminal states from the desired position $\bm{x}_d$ and velocity $\bm{v}_d$:
\begin{equation}
    \Gamma_{x_d} = w_{x_d} \sum_{k = \mathbb{K}_f} \left(\|\bm{x}(k) - \bm{x_d}\|^2 + \|\bm{v}(k) - \bm{v}_d\|^2\right)
\end{equation}
where $w_{x_d} \in \mathbb{R}^+$ is a tunable gain. $\mathbb{K}_f$ contains a certain portion of the last phase, including the terminal state. Thus, it is possible to generate trajectories which reach the end point in advance while defining control inputs able to maintain such position. 

The model defined by Eq. \eqref{eq:double_pendulum} does not carry any information about the joint torques necessary to achieve the planned motion. Nevertheless, with the aim of reducing the torques required at the leading knee to step-up, we introduce the following task:
\begin{equation*}
    \Gamma_\tau = w_\tau \sum_{k,*} \uptau_*(k)^2 + w_{\uptau_\text{max}}\sum_{*}\left( \max_k \uptau_*(k)^2  \right),
\end{equation*}
where
\begin{equation}
    \uptau_*(k) = \begin{cases}
                        \left(x_z(k) - x_{*,z} - \delta_* \right) \lambda_*(k) & \text{if $*$ in contact}\\
                        0 & \text{otherwise},
                  \end{cases}
\end{equation}
is used as a heuristic to reduce joint torques. $*$ is again a placeholder for $l$ and $r$, while $\delta_*$ is a reference height difference. Intuitively, the robot knees undergo higher stress when they are highly bent. In this configuration, the height difference between the CoM and the foot gets small. This heuristic prevents the solver from requiring high forces (characterized by a high multiplier $\lambda_*$) on a collapsed leg. Thus $\Gamma_\tau$ weights the summation over each time instant of this quantity squared. The second term is employed to prevent this quantity to have an impulsive behavior. $w_\tau \in \mathbb{R}^+$ and $w_{\uptau_\text{max}} \in \mathbb{R}^+$ are weights allowing to specify the relative priority for each task.

Considering the control inputs $\lambda_*(k)$ and $\bm{p}_*(k)$, it is preferable to generate trajectories which require small control variations. This reduces the need for large torque variations when tracking the trajectories on the real robot. To this end, we consider the following task:
\begin{equation}
    \Gamma_{\Delta_u} = w_{\Delta_u} \sum_{k=2\dots N \cdot P} \|\bm{u}_*(k) - \bm{u}_*(k-1)\|^2,
\end{equation}
where $w_{\Delta_u} \in \mathbb{R}^+$ is the task weight and $\bm{u}_* = \left[\lambda_*^\top~\bm{p}_*^\top\right]^\top$.

Finally, we consider a series of regularization terms: 
\begin{equation}
\begin{split}
    \Gamma_\text{reg} =& w_t \sum_i \|T_i - T_{i, d}\|^2 + w_\lambda \sum_k \|\lambda_*(k)\|^2 +\\
    &+ w_p \sum_k \|\bm{p}_*(k)\|^2.
\end{split}
\end{equation}
This task allows selecting solutions which minimize the use of control inputs and the duration of each phase is close to a desired one $T_{i, d}$. $w_t, w_\lambda, w_p \in \mathbb{R}$ are the tasks weights.

\subsection{The optimization problem}
The trajectory optimization problem constituted by the dynamic constraint defined in Eq. \eqref{eq:discrete_double_pendulum}, the constraints listed in Sec. \ref{sec:force_constraints} and the tasks of Sec. \ref{sec:double_pendulum_tasks}, is casted into an optimization problem via a Direct Multiple Shooting method \cite{betts2010practical}. The optimization variables $\bm{\chi}$ correspond to the following set:
\begin{equation*}
\bm{\chi} = \begin{cases}
\bm{x}(k) &k = 0 \dots N \cdot P \\
\bm{v}(k) &k = 0 \dots N \cdot P \\
\bm{a}(k) &k=1 \dots N \cdot P \\
\lambda_*(k) &k=1 \dots N \cdot P,~ *=l,~ r \\
\bm{p}_*(k) &k=1 \dots N \cdot P,~ *=l,~ r  \\
T_i &i=1\dots P,
\end{cases}
\end{equation*}
with $P \in \mathbb{N}^+$ equal to the number of phases. $\bm{x}(k)$ and $\bm{v}(k)$ are state variables and need to be initialized with the measurements coming from the robot $\bm{x}_0$, $\bm{v}_0$, i.e. $\bm{x}(0) = \bm{x}_0$ and $\bm{v}(0) = \bm{v}_0$.

The complete formulation is shown in Optimization Problem \ref{alg:step-up-planner}. Notice that the subscript $i$ refers to quantities which depend on a specific phase and they are fixed for its entire duration. It is implemented using \texttt{CasADi} \cite{Andersson2018}, adopting \texttt{Ipopt} \cite{IPOpt2006} as solver.
\begin{algorithm}
\SetKwBlock{SubjectTo}{subject to:}{}
\SetInd{0.15em}{0.6em}
 \begin{flalign*}
     \minimizebf_\chi & ~~\Gamma_{x_d} + \Gamma_\tau + \Gamma_{\Delta_u} + \Gamma_\text{reg}&&
 \end{flalign*}
 \SetAlgoLined
 \SubjectTo{
    $\bm{x}(0) = \bm{x}_0$ \\
    $\bm{v}(0) = \bm{v}_0$ \\
  \For{$i = 1 \dots P$} {
    $T_{i}^\text{ min} \leq T_i \leq T_{i}^\text{ max}$ \\
    $\mathrm{dt}_i := T_i/N$\\
    \For{$k = \left((i-1)P\right)$ $\dots$ $\left(i \cdot P - 1 \right) $ } 
    {
        $\bm{x}(k+1) = \bm{x}(k) + \mathrm{dt}_i ~ \bm{v}(k) + \frac{1}{2}\mathrm{dt}_i^2~\bm{a}(k)$\\
        $\bm{v}(k+1) = \bm{v}(k) + \mathrm{dt}_i~\bm{a}(k)$\\
        $\bm{\mathrm{a}} := -\bm{g}$\\
        \If{left in contact} {
            $\bm{\mathrm{a}} := \bm{\mathrm{a}} + \lambda_l(k) \left(\bm{x}(k){-} \bm{x}_{l, i} {-} \bm{R}_{l, i} \bm{p}_l(k) \right)$\\
            $\bm{A}_{l, i}~\bm{p}_l(k) \leq \bm{b}_{l, i}$\\[2pt]
            $\left[1 ~ 1 ~ {-}\mu^2\right] \left(\bm{R}_{l, i}^\top \left(\bm{x}(k) {-} \bm{x}_{l, i} {-} \bm{R}_{l, i} \bm{p}_l(k) \right)\right)^2 {\leq} 0$\\[2pt]
            $\bm{F}_{l,i} \bm{R}_{l,i}^\top \left(\bm{x}(k) - \bm{x}_{l,i} - \bm{R}_{l,i} \bm{p}_l(k) \right) \leq 0$\\[2pt]
            $l^\text{ min} \leq \|\bm{x} - \bm{x}_{l, i}\| \leq l^\text{ max}$\\
        }
        \If{right in contact} {
            $\bm{\mathrm{a}} := \bm{\mathrm{a}} + \lambda_r(k) \left(\bm{x}(k){-} \bm{x}_{r, i} {-} \bm{R}_{r, i} \bm{p}_r(k) \right)$\\
            $\bm{A}_{r, i}~\bm{p}_r(k) \leq \bm{b}_{r, i}$\\[2pt]
            $\left[1 ~ 1 ~ {-}\mu^2\right] \left(\bm{R}_{r, i}^\top \left(\bm{x}(k) {-} \bm{x}_{r, i} {-} \bm{R}_{r, i} \bm{p}_r(k) \right)\right)^2 {\leq} 0$\\[2pt]
            $\bm{F}_{r,i} \bm{R}_{r,i}^\top \left(\bm{x}(k) - \bm{x}_{r,i} - \bm{R}_{r,i} \bm{p}_r(k) \right) \leq 0$\\[2pt]
            $l^\text{ min} \leq \|\bm{x} - \bm{x}_{r, i}\| \leq l^\text{max}$\\
        }
        $\bm{a}(k) = \bm{\mathrm{a}}$
    }
  }
 }
 \SetAlgorithmName{Optimization Problem}{ }
 
 \caption{ } 
 \label{alg:step-up-planner}
\end{algorithm}

The optimization problem is non-convex due to the non-linearities of the model equation, of the friction constraints and the chosen tasks. In order to facilitate the finding of a solution, we initialize the phases duration $T_i$ with their desired value. The CoM position trajectory is initialized to a linear interpolation from $\bm{x}_0$ to $\bm{x}_d$. All other variables are initialized to zero.

\section{Validation} \label{sec_ls:results}
\begin{figure*}[tpb]
    \centering
    \subfloat[$t=3.3s$] {\includegraphics[width=.22\textwidth]{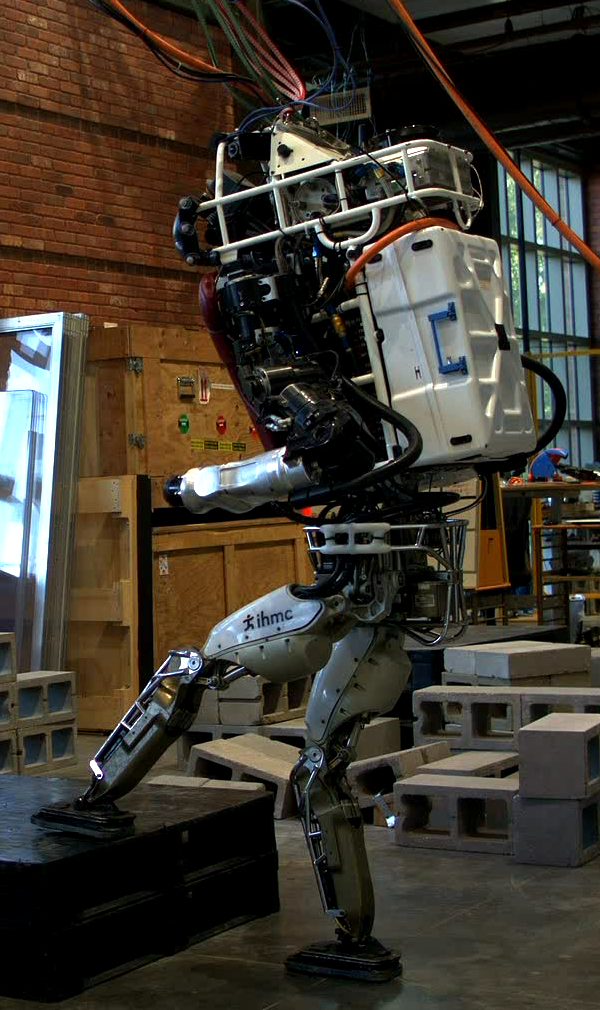}}
    \hspace{.01\textwidth}
    \subfloat[$t=4.2s$] {\includegraphics[width=.22\textwidth]{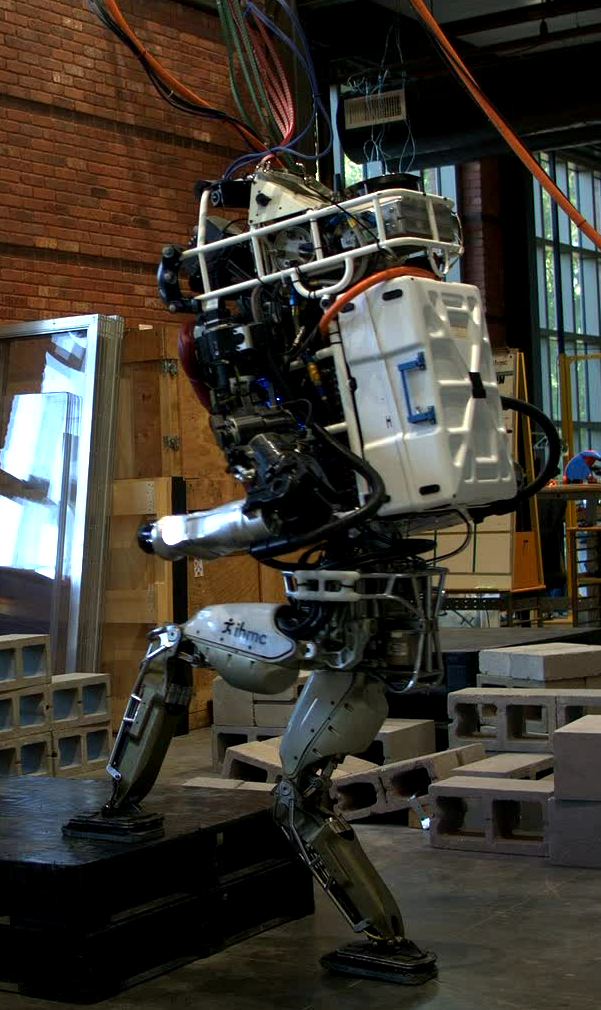}\label{fig_ls:lowestCoM}}
    \hspace{.01\textwidth}
    \subfloat[$t=5.6s$] {\includegraphics[width=.22\textwidth]{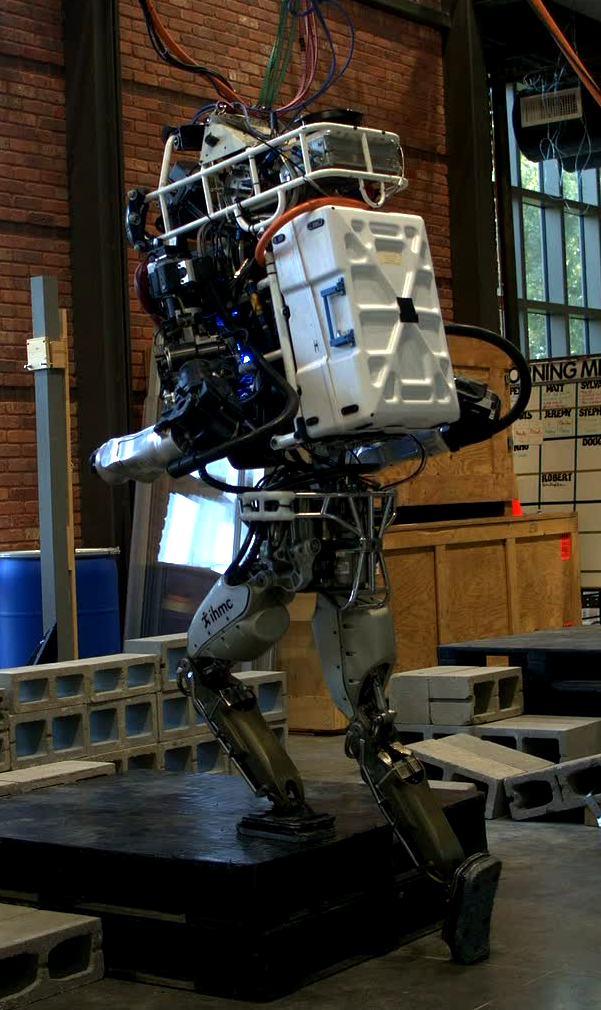} \label{fig_ls:endOfDoubleSupport}}
    \hspace{.01\textwidth}
    \subfloat[$t=8.8s$] {\includegraphics[width=.22\textwidth]{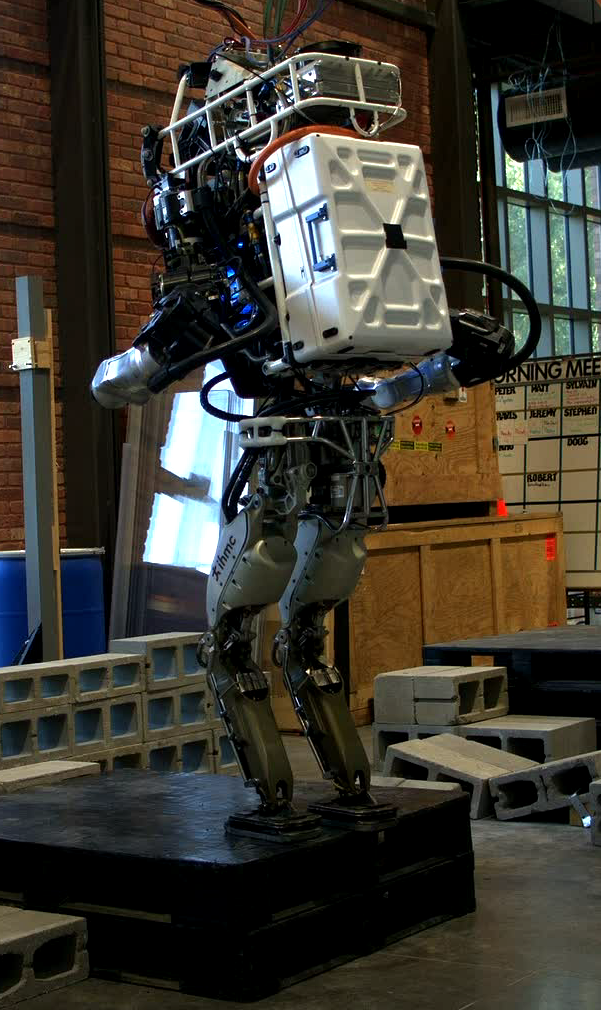}}
    \caption{Snapshots of the step-up motion. They correspond to switches of phases, except for the one at time $4.2s$ which corresponds to the lowest CoM height.}
    \label{fig_ls:snapshots}
\end{figure*}
 \begin{figure}[tpb]
 	\centering
 	\subfloat[$x$ direction.]{ 	\includegraphics[width=0.9\columnwidth]{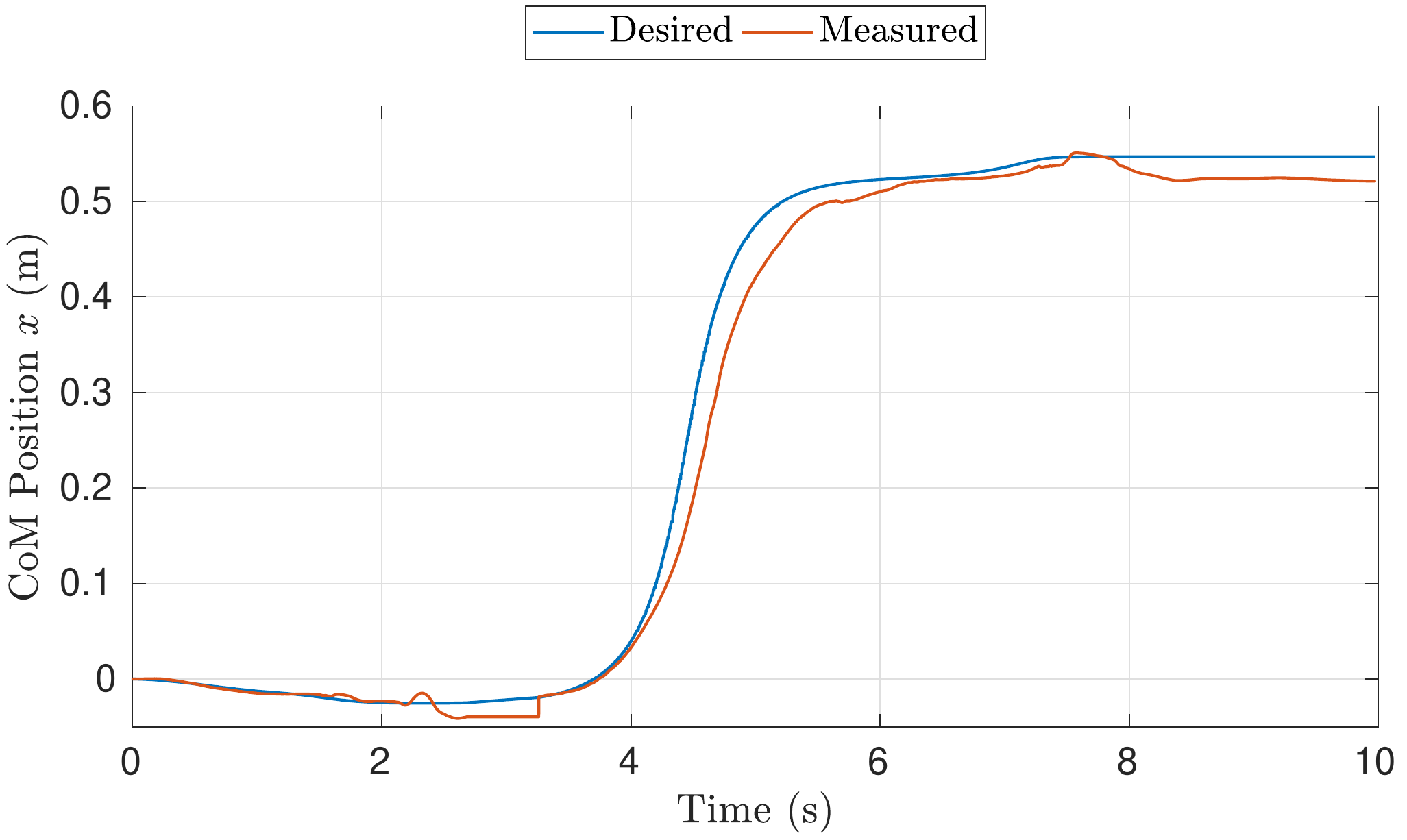} \label{fig_ls:jump_comx}}
 	
 	\subfloat[$y$ direction.]{ 	\includegraphics[width=0.9\columnwidth]{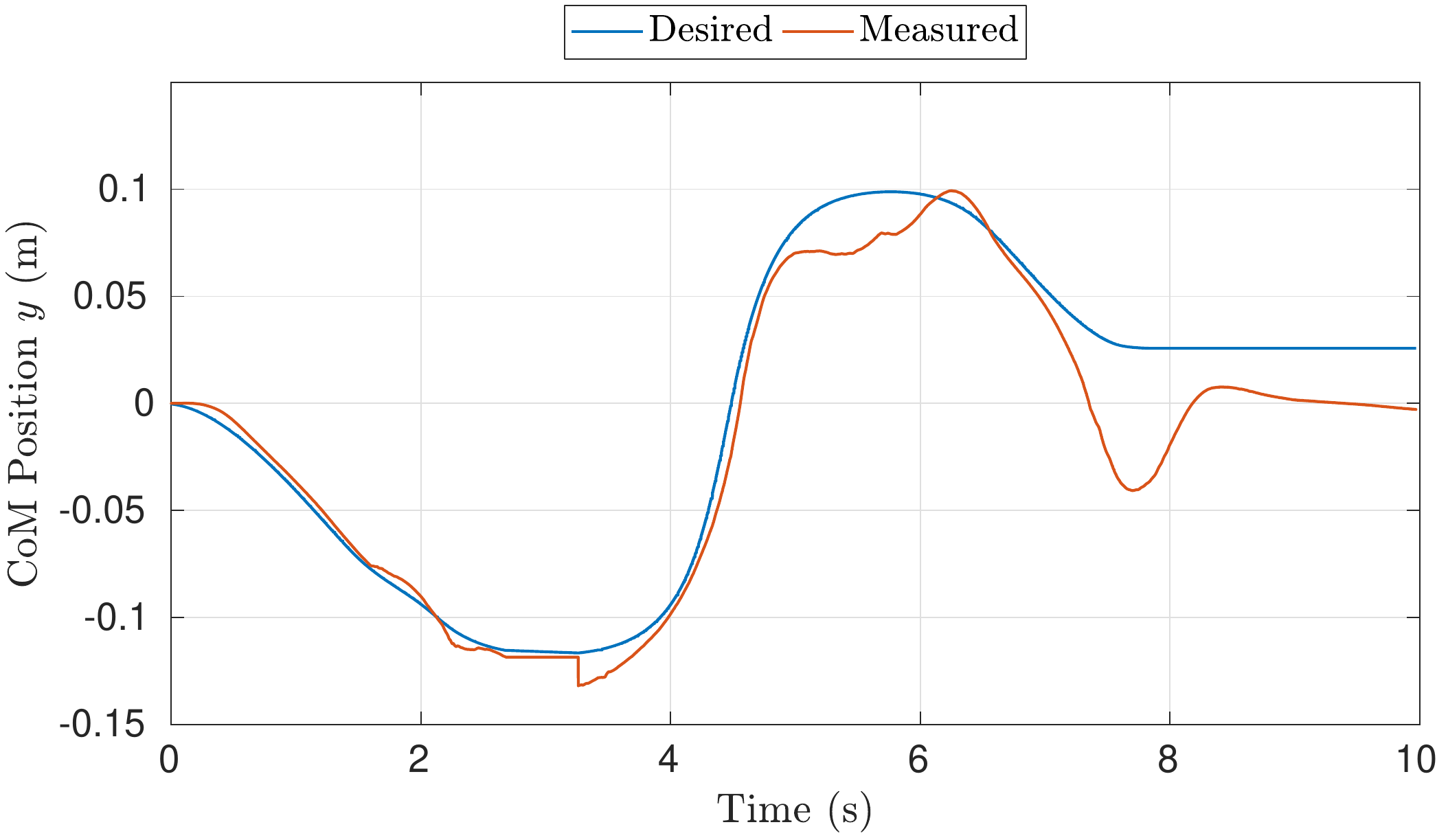}  \label{fig_ls:jump_comy}}
 	
 	\subfloat[$z$ direction.]{ 	\includegraphics[width=0.9\columnwidth]{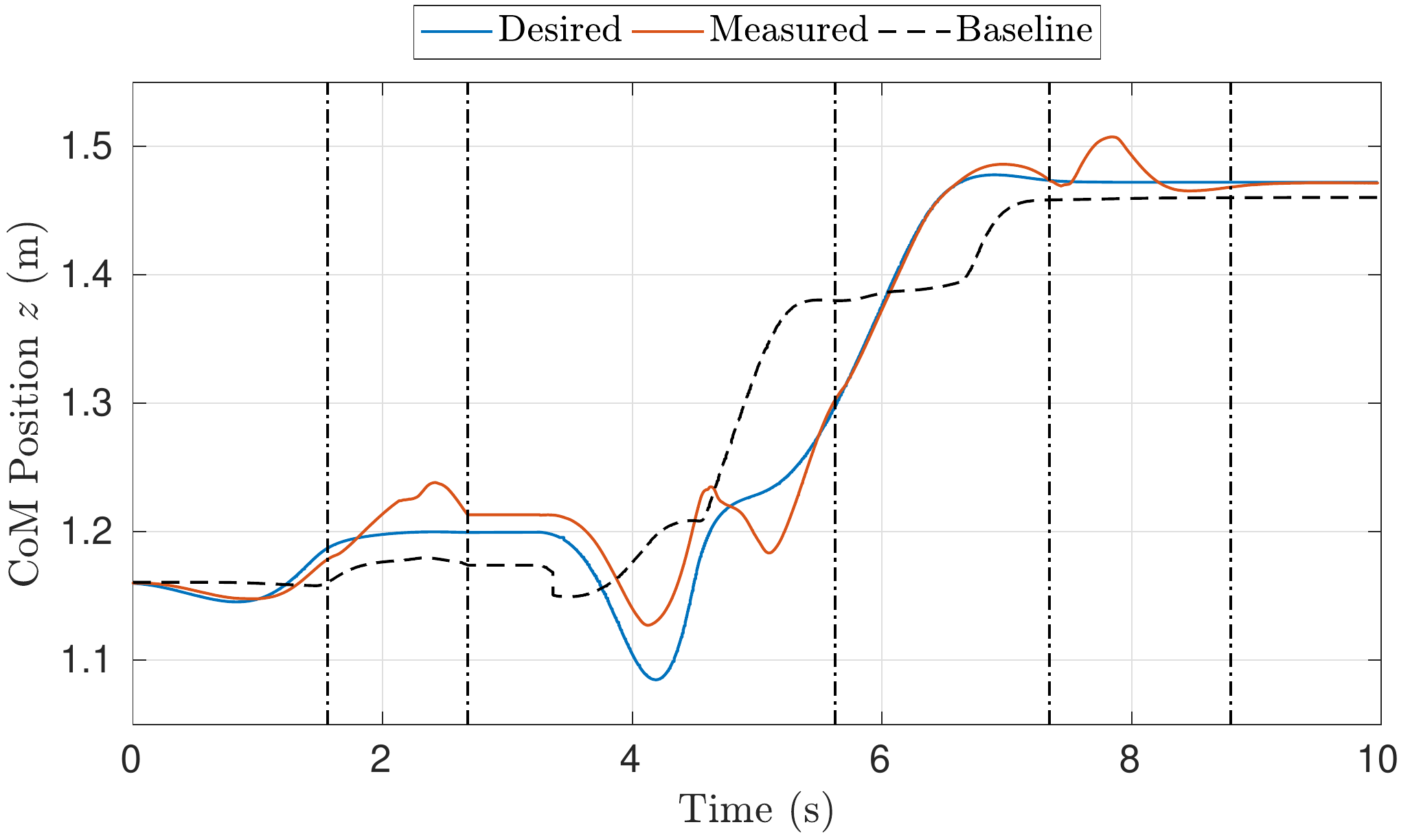}  	\label{fig_ls:jump_comz}}
 	
 	\caption{Tracking of the CoM position. Measured quantities are obtained through an internal estimator.}
 	
 \end{figure}
\begin{figure}[tpb]
    \centering
    \includegraphics[width=0.9\columnwidth]{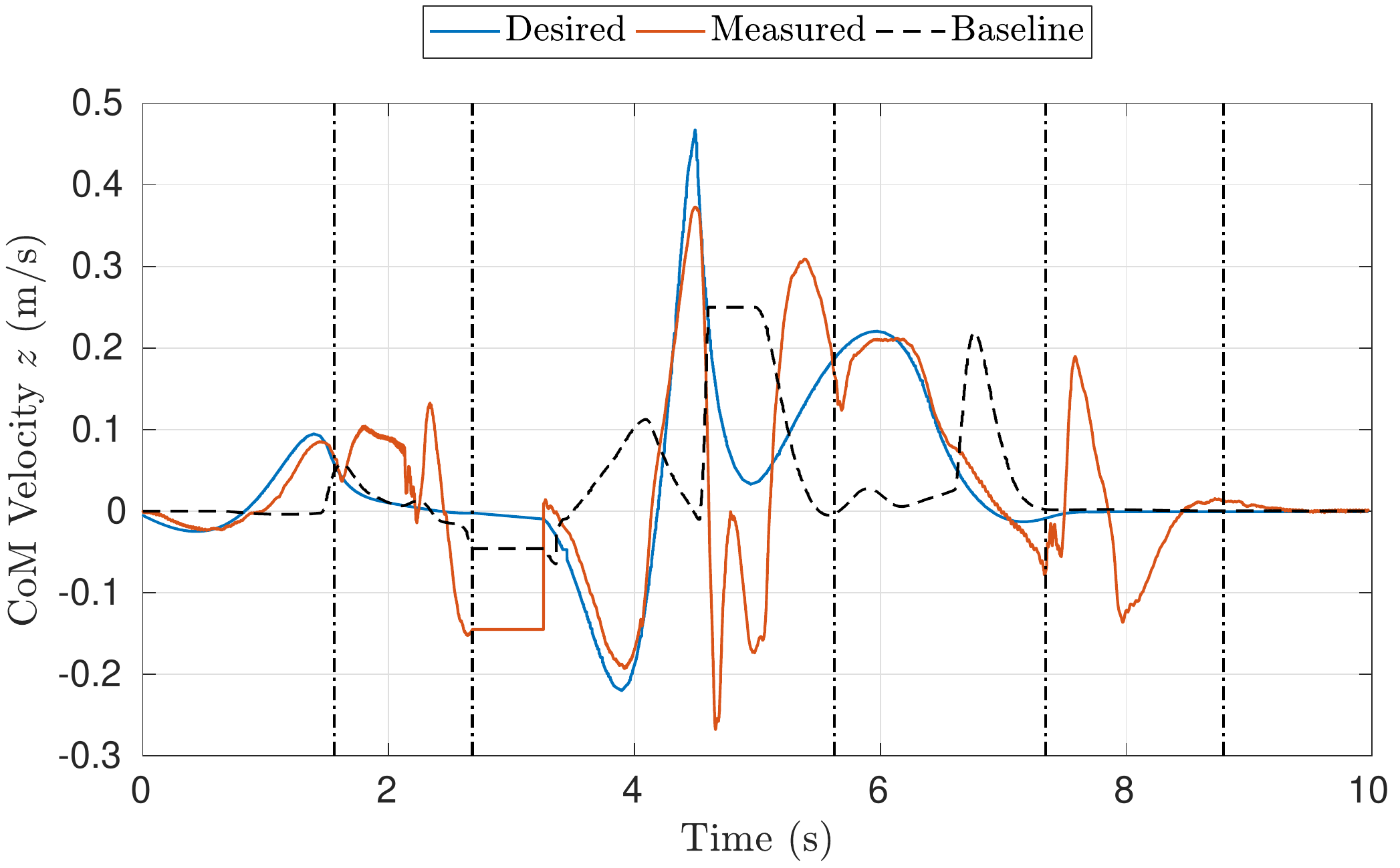}
    \caption{Tracking of the desired height velocity. The vertical dotted lines indicate the change of phases. It is possible to notice that the tracking worsens around $t=5s$, which is close to the end of the double support phase.}
    \label{fig_ls:jump_comVelocityz}
\end{figure}
\begin{figure}[tpb]
    \centering
    \includegraphics[width=0.9\columnwidth]{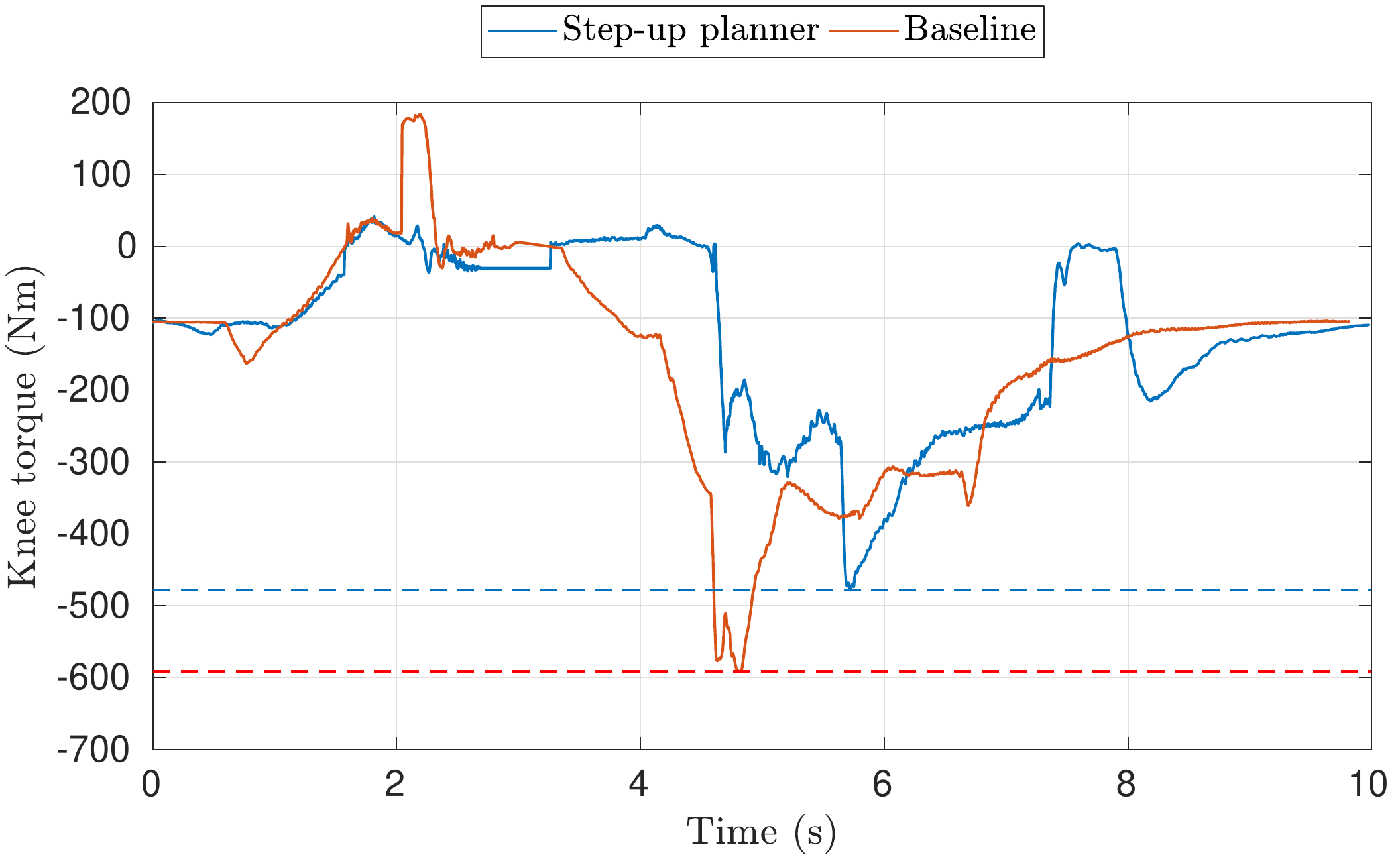}
    \caption{Comparison of measured knee torques. The baseline is obtained by letting the controller described in \cite{koolen2016balance} generating the motion automatically. 
    The dotted lines are the peak torques. The maximum torque required to the robot when using the step-up planner is $20\%$ lower than the baseline.}
    \label{fig_ls:jump_torques}
\end{figure}

The trajectories generated with the method described in Sec. \ref{sec_ls:sup} are stabilized by the controller presented in Sec. \ref{sec_ls:qp_controller}. The desired accelerations and contact wrenches are turned into a set of desired joint torques commanded to the Atlas humanoid robot.

The task consists in stepping onto a platform whose height is about 31$\mathrm{cm}$, as showed in Figure \ref{fig:atlasShowingOff}. We noticed that this height is already enough to have the robot pushing the leg joint limits during the swing phase. The robot starts at a distance of about 30$\mathrm{cm}$ from the platform and performs a 55$\mathrm{cm}$ step forward on the platform. Once the robot is close to the platform, the optimization is started using the current robot state as initialization. The optimization problem is solved using a desktop computer equipped with a 5$^{th}$ generation Intel\textsuperscript{\textregistered} Core i7@3.3GHz processor and 64GB RAM memory, running Ubuntu 16.04. \texttt{Ipopt} is set up to use the \texttt{mumps} \cite{amestoy2000mumps} solver. A solution is generated after about 1.5$\mathrm{s}$. This time can be reduced up to a factor 3 by using the linear solver \texttt{ma27} \cite{hsl2007collection} instead of \texttt{mumps}. Nevertheless, considering that the trajectories are generated only once before the beginning of the motion, this improvement is not necessary at this stage.

The length of each phase is set to $N=30$ instants. We noticed that this value is a good trade-off between the computational time and the smoothness of the generated trajectories. The maximum duration of a phase is equal to $2.3\mathrm{s}$, thus corresponding to a maximum $\mathrm{dt} = 77\mathrm{ms}$.

Amongst the set of variables $\bm{\chi}$ constituting the solution provided by the solver, we pack the CoM state into a desired trajectory. In addition, the timings $T_i$ are used to determine the single and double support phase durations. 
\subsection{Real robot experiments}
Fig. \ref{fig_ls:snapshots} contains snapshots of a step-up executed by the Atlas humanoid robot when using the planner. Fig. \ref{fig_ls:jump_comx} and \ref{fig_ls:jump_comy} show the corresponding CoM position tracking along the $x$ and $y$ direction. Figure \ref{fig_ls:jump_comz} presents the tracking of the CoM height. Note that the CoM is lowered right before performing the step-up. This instant is shown in Fig. \ref{fig_ls:lowestCoM}. We believe such desired motion arises to achieve sufficiently high vertical velocity, see Fig. \ref{fig_ls:jump_comVelocityz}, while considering the limits on the leg length. By performing this motion, the robot gains momentum to perform the step-up motion requiring a lower torque on the leading knee.

In order to provide a comparison, we performed the same task without prescribing any CoM trajectory, thus adopting the motion generation techniques presented in \cite{koolen2016design}. To facilitate the comparison, we impose the same phases duration. Fig. \ref{fig_ls:jump_torques} shows the torque profile measured on the left knee when performing the step-up. We focus on this joint since is the one witnessing the highest torque expenditure during the step-up motion. In particular, using the controller presented in \cite{koolen2016balance}, the maximum value of this torque reaches $591\mathrm{Nm}$, while it is reduced to $478\mathrm{Nm}$ when using the method presented in this paper. This corresponds to a nearly $20\%$ maximal torque reduction. Such result comes at the cost of a higher torque on the trailing knee. In order to facilitate the step-up motion, the right leg is much more exploited compared to the baseline, with a knee torque of nearly $430\mathrm{Nm}$.

Since the trajectories are computed offline, small disturbances can induce the robot to a fall. Robustness is achieved by limiting the CoP to a smaller region compared to the real robot foot, in the planning phase. In this way there is margin in the underlying QP controller for counteracting disturbances while executing those trajectories. During the experiment described above, the foot dimension is set to $30\%$ of the original size. 

\begin{figure}[tpb]
 	\centering
 	\includegraphics[width=0.9\columnwidth]{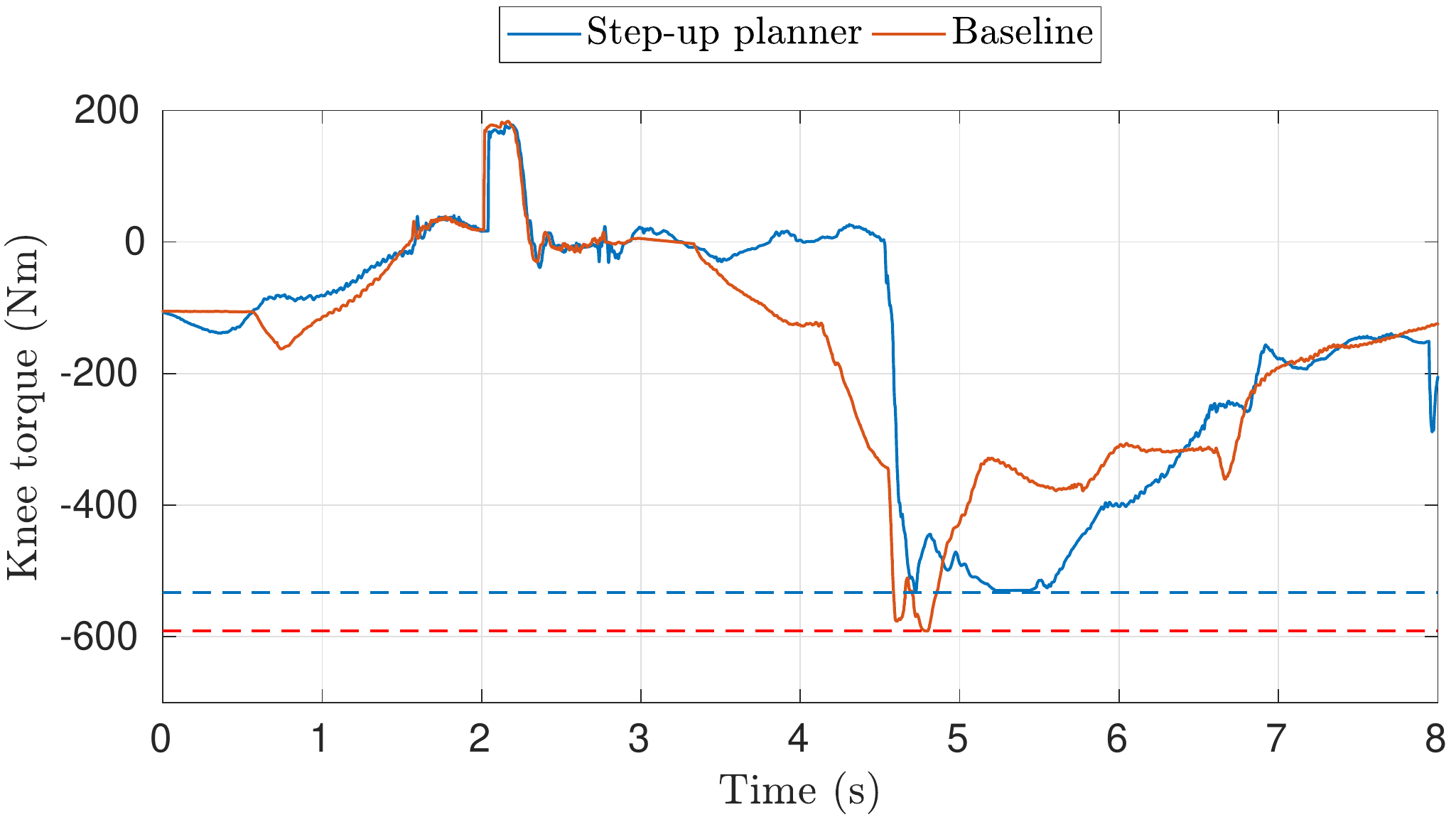}
 	\caption{Leading knee torques when the static friction is reduced to 0.5. Here the reduction of the maximum value is about 10\%.}
 	\label{fig_ls:reliable_torques}
 \end{figure}

The planned step-up is a dynamic motion which requires a high vertical velocity of the CoM. We noticed that the static friction coefficient $\mu_s$ has a particular effect in determining how ``dynamic'' is the planned motion. Intuitively, it limits the angle between the normal to the ground and the pendulum. A small angle corresponds to a conservative motion where the projection of the CoM on the ground lies well inside the support polygon. In the experiment shown above, the static friction coefficient is set to 0.7. By reducing it to $0.5$ the motion is more conservative and robust, even though the effectiveness of the method is reduced to a maximum torque reduction of 10\%. As shown in Fig. \ref{fig_ls:reliable_torques}, the knee torque reaches a peak of $532\mathrm{Nm}$.

 \begin{figure}[tpb]
 	\centering
 	\subfloat[$x$ direction.]{ 	\includegraphics[width=0.9\columnwidth]{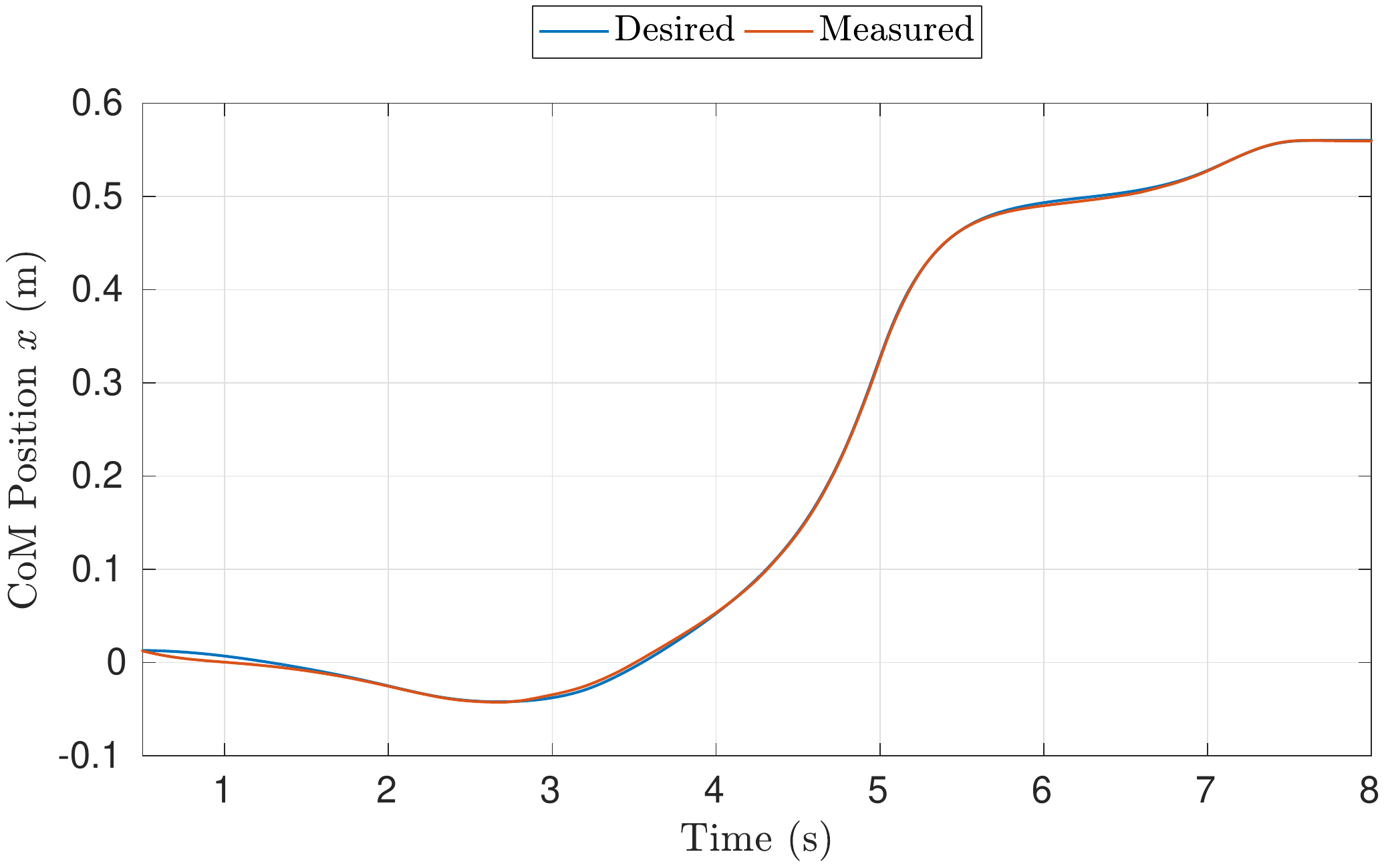} \label{fig_ls:sim_comx}}
 	
 	\subfloat[$y$ direction.]{ 	\includegraphics[width=0.9\columnwidth]{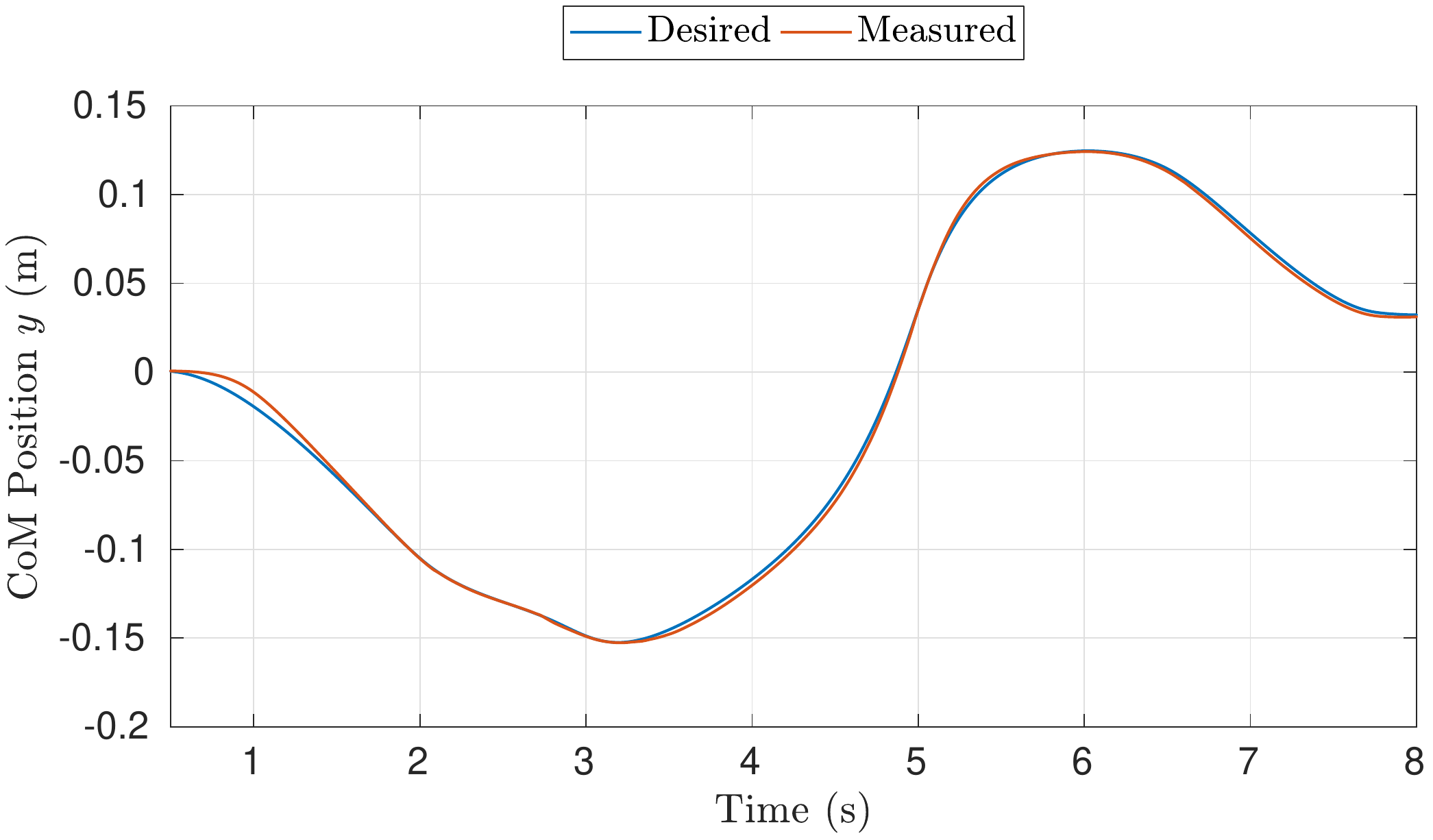}  \label{fig_ls:sim_comy}}
 	
 	\subfloat[$z$ direction.]{ 	\includegraphics[width=0.9\columnwidth]{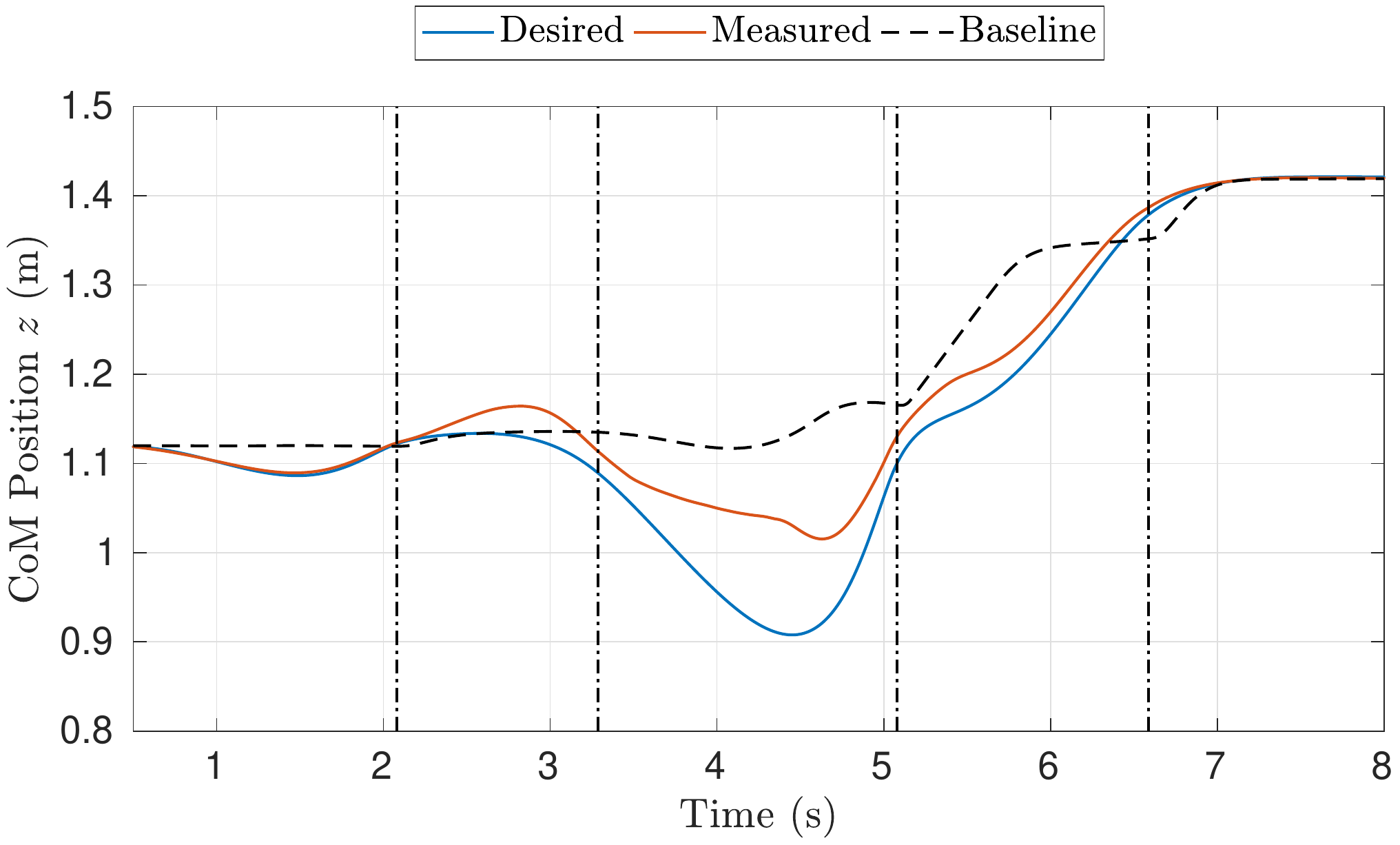}  	\label{fig_ls:sim_comz}}

 	\caption{CoM position tracking in the simulation experiment.}
 	
 \end{figure}
 \begin{figure}[tpb]
 	\centering
 	\includegraphics[width=0.9\columnwidth]{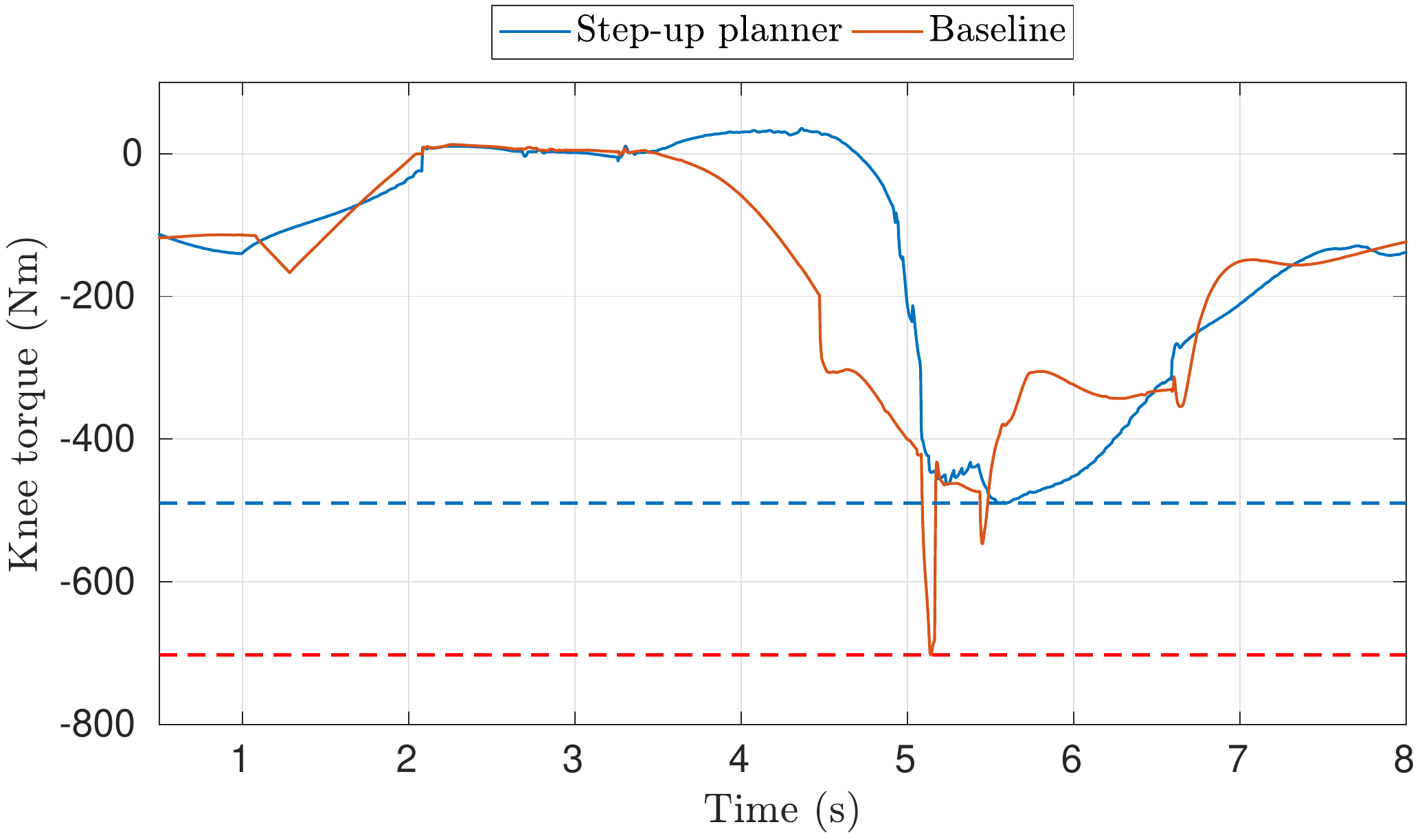}
 	\caption{Comparison of measured knee torques during the simulation experiment.}
 	\label{fig_ls:sim_torques}
 \end{figure}

When performing experiments on the real robot, the poor tracking of the desired trajectories limits the robustness and reproducibility of the results. Since the trajectories are computed offline, a poor tracking limits their efficacy. In the majority of the experiments, the reduction of the maximum torque consisted of about $10\%$. Anyhow, we believe this still represents an interesting achievement. It shows that trajectory optimization techniques have an effect in reducing the maximum effort required to a $150\mathrm{kg}$ robot when performing a motion at the limits of its reachable workspace.

\subsection{Simulation results}
When performing the same test in simulation, the tracking on the planar directions is much better, as shown in Fig. \ref{fig_ls:sim_comx} and \ref{fig_ls:sim_comy}, while along the $z$ direction, it is somehow comparable, Fig. \ref{fig_ls:sim_comz}. The improved tracking helps in maintaining balance, allowing more dynamic motions. The figures shown are generated with a static friction coefficient equal to 1.0, while the timings are kept equal to the real robot experiments. The baseline is obtained in the same way, leaving to the controller the generation of desired trajectories given a defined sequence of footsteps. As it can be noticed from Fig. \ref{fig_ls:sim_torques}, when the robot switches from double to single support the knee torque has a spike of $702\mathrm{Nm}$. Given the coincidence with the change of support configuration, the source of such spike is most probably the controller itself. Since in simulation the torque control is almost perfect, this spike is reflected also in the measured data. Conversely, on the real robot, any spike on the desired torques is smoothed due to the actuator dynamics. Nevertheless, when using the planner there is no spike and the maximum torque is at $489\mathrm{Nm}$, corresponding to a $30\%$ maximum torque reduction.

\begin{table}[tpb]
	\caption{Cost function weights used both in simulation and in the real robot experiments. In the experiments presented, $N=30$ while $P=5$.}
	\centering
	\begin{tabular}{|c c c c c c c|} 
		\hline
		$w_{x_d}$ & $w_\tau$ & $w_{\uptau_\text{max}}$ & $w_{\Delta_u}$ & $w_t$ & $w_\lambda$ & $w_p$\Tstrut\Bstrut\\ [0.5ex] 
		\hline\hline
		$10.0$ & $\frac{0.1}{N\cdot P}$ & $4.0$ & $\frac{200.0}{N\cdot P}$ & $\frac{5.0}{P}$ & $\frac{0.1}{N\cdot P}$ & $\frac{1.0}{N\cdot P}$\Tstrut\Bstrut\\ [1ex] 
		\hline
	\end{tabular}
	\label{tab:weights}
\end{table}
Table \ref{tab:weights} provides the weights adopted in the cost function tasks of Sec. \ref{sec:double_pendulum_tasks}. They are the same in simulation and in real robot experiments. The tuning process consists in adding one task at a time starting from the most important, i.e. $\Gamma_{x_d}$, which is assigned an arbitrary weight. It is possible to notice that the three highest weights are those related to the terminal state tracking, the maximum $\uptau_*$ and the control variations.
\section{Conclusions} \label{sec_ls:conclusions}
This paper presents a method to generate trajectories for large step-up motions with humanoid robots. Its effectiveness has been tested on the IHMC Atlas humanoid robot. Experiments carried on the real platform showed that such trajectories can reduce the maximum torque required to the knee up to $20\%$. 

The method has been studied for large step-ups, but its applicability is not limited to this domain. Indeed, its formulation enables the planning of motions involving flight phases, hence requiring dynamic movements. This represents a fascinating future work.

The desired trajectories are computed offline right before starting the step-up motion. This reduces the robustness of the planned trajectories, strongly relying on the tracking performances of the low-level controller. Poor tracking of the planned trajectories limits also their efficacy in reducing the maximum knee torque.
A possible future work consists in re-planning the trajectories every time a walking phase is completed. However, the computational time would need to be reduced by at least an order of magnitude.

\addtolength{\textheight}{0cm}   

\addcontentsline{toc}{section}{References}

\bibliography{IEEEabrv,Bibliography}

\end{document}